# An Evolutionary Correlation-aware Feature Selection Method for Classification Problems

Motahare Namakin[1], Modjtaba Rouhani[1,*] and Mostafa Sabzekar[2]

1- Department of Computer Engineering, Ferdowsi University of Mashhad, Mashhad, Iran
2- Department of Computer Engineering, Birjand University of Technology, Birjand, Iran

**Abstract-** As global search techniques, population-based optimization algorithms have provided promising results in feature selection (FS) problems. However, the main challenges are high time complexity due to the exploration of a large search space and consequently a large number of fitness function evaluations. Moreover, the interaction between features is another big challenge in FS problems that directly affects the classification performance by selecting correlated features. In this paper, an estimation of distribution algorithm (EDA)-based method is proposed to meet three goals. Firstly, as an extension of EDA, the proposed method generates only two individuals in each iteration that compete based on a fitness function and evolve during the algorithm, based on our proposed update procedure. Secondly, we provide a guiding technique for determining the number of features for individuals in each iteration. As a result, the number of selected features of the final solution will be optimized during the evolution process. The two mentioned advantages can increase the convergence speed of the algorithm. Thirdly, as the main contribution of the paper, in addition to considering the importance of each feature alone, the proposed method can consider the interaction between features. Thus, it can deal with complementary features and consequently increase classification performance. To do this, we provide a conditional probability scheme that considers the joint probability distribution of selecting two features. The introduced probabilities successfully detect correlated features. Experimental results on a synthetic dataset with correlated features prove the performance of our proposed approach facing these types of features. Furthermore, the results on 13 real-world datasets obtained from the UCI repository show the superiority of the proposed method in comparison with some state-of-the-art approaches. The efficiency analysis of the experimental results using two non-parametric statistical tests prove that the proposed method has significant advantages in comparison to other approaches.

## 1. Introduction

Classification as one of the major tasks of machine learning has been remarkably applied to vast domains of research such as Bioinformatics [1], intrusion detection systems (IDSs) [2], fraud detection [3], and prediction of different diseases (Parkinson [4], Cancers [5], COVID-19 [6], etc.). One of the critical preprocessing steps in classification applications is feature selection (FS). To improve the classification performance and build a robust model, it is essential to search and find the optimal subset of features such that the selected features be as informative and small as possible. Thus, the main goals of FS can be summarized as reducing the computational time and cost of building the model, preventing over-fitting and increasing the generalizability of the obtained classifier, and improving its accuracy by removing redundant and irrelevant features. However, without prior knowledge, it is difficult to discriminate useful features from others. Moreover, FS is a challenging research problem not only due to the large search space, but also because of the correlation of features. It is clear that the size of search space increases exponentially with the number of features. It makes the FS problem NP-hard, and therefore, exhaustive search is an impractical solution. Besides, any correlation between features can considerably decrease the classification accuracy.

---

* Corresponding author

There are many studies in the literature that aims to overcome those FS challenges. Based on the evaluation criteria, these efforts can be generally classified into: *filter*, *wrapper*, and *embedded* approaches [7]. Filter feature selection methods are based merely on the inherent characteristics of data. They generate candidate feature subsets without involving any classification algorithm in evaluating phase. The wrapper approaches utilize a predetermined learning algorithm and consider their performance as the goodness criterion of feature subsets. Therefore, they generally achieve better performances than filter-based methods. However, these algorithms are computationally expensive and need more time to execute compared to filter-based algorithms. Finally, the methods incorporating feature selection and training process of the learning model into a single procedure are called embedded approaches. Embedded approaches combine the advantages of two other categories to make a good trade-off between computation time and accuracy. The main limitation of this category is that it can be applied to specific learning models.

From the viewpoint of the searching techniques, the FS methods can be categorized into sequential and global search methods [8]. The most common methods in the former category are the sequential forward selection (SFS) and the sequential backward selection (SBS). The SFS (SBS) starts with the empty (full) set and progressively adds (removes) features until the performance of the classifier is improved (not decreased). However, the sequential-based methods are typically stuck in local optima. Additionally, the search strategy of the latter category is based on the random search in the solution space to find the best feature subset. Nowadays, population-based optimization approaches, as global search methods, report promising and successful results dealing with the FS problem [9]. The most common population-based optimization methods in solving the FS problem include genetic algorithm (GA) [10]–[12], particle swarm optimization (PSO) [13]–[16], ant colony optimization (ACO) [17]–[19], artificial bee colony (ABC) [20]–[22], cuckoo search (CS) [23], [24]. The main advantages of population-based optimization approaches are, firstly, they do not need domain knowledge or any assumption about the problem, and secondly, they can generate several solutions in a single run due to their population-based structure [8]. However, the main challenge is that they suffer from high time complexity because of the exploration of a large search space and consequently a large number of function evaluations. Another limitation of all FS methods is that they usually cannot consider the interaction between features. On the one hand, a feature may be weakly relevant to the class label individually, but the classification accuracy can be improved by combining it with some complementary features. On the other hand, a feature may be relevant by itself, but it may also decrease the classification accuracy or generalization when used with some other correlated features.

To overcome the limitations of feature selection methods that utilized population-based optimization, our main contribution is to present a new method based on the estimation of distribution algorithm (EDA) that considers the correlated features. EDA [25], [26] is one of the population-based optimization methods that generate the new candidate solutions by a probabilistic model. The optimization in this method is considered as a sequence of probabilistic model updates. Although in conventional population-based optimization methods, the new candidate solutions are generated utilizing an implicit distribution (variation operators), the EDA utilizes an explicit probability distribution such as multivariate interaction, Bayesian network, etc.

Consequently, in this paper, we present a correlation-aware feature selection approach so that in addition to considering the importance of each feature alone, it can also consider the interaction between features. Thus, the proposed method is able to deal with complementary features and consequently improves the classification performance. For this purpose, we introduce a conditional probability scheme that considers the joint probability distribution of selecting two features. Furthermore, as mentioned before, one of the main drawbacks of the wrapper-based FS methods, which utilized evolutionary algorithms, is that they suffer from high time complexity due to a large number of function evaluations. Nevertheless, the proposed method generates only two individuals in each iteration that compete based on a fitness function and evolve during the algorithm execution. Thus, we considerably reduce the overhead of calculating the number of function evaluations for wrapper-based FS methods. As a result, the proposed approach will be quite fast for solving FS problems. Finally, we propose a guiding technique that helps the algorithm to select the appropriate number of features during the evolution process.

The rest of this paper is organized as follows. Section 2 provides a brief review of the state-of-the-art FS methods, which utilized population-based optimization algorithms. The architecture of the proposed method is illustrated in Section 3. Finally, the experimental results followed by some conclusions and future works are discussed in Sections 4 and 5, respectively.

## 2. Literature review

As mentioned before, the population-based optimization approaches for solving the FS problem have been widely used in the literature. There are many ways that we can categorize and study these efforts. Here, we study them from two aspects which are the representation method and the number of objectives. Based on the representation method, the FS algorithms that utilized population-based optimization can be classified into binary and continuous representation. In binary representation, 1s and 0s are used as values of each vector element of individuals, which indicates selecting or non-selecting features. On the other hand, the continuous representation consists of real values. Generally, in continuous representation, a threshold $\theta$ is considered to determine that the corresponding feature should be selected or discarded. If the element value for a feature is greater than $\theta$, it will be selected by the FS method, and otherwise, it will be dropped.

Besides, based on the number of objectives, we can categorize them into single-objective (SO) and multi-objective (MO) algorithms. The SO methods usually consider only the classification accuracy as the objective function. In contrast, MO methods apply other criteria as well. For instance, in many MO methods, the number of selected features is considered as the second objective function in addition to the primary accuracy of the classification algorithm. There are many approaches in each category that can be reviewed. In the following this section, we survey only the state-of-the-art researches and compare their capabilities.

Genetic algorithm is the most common and likely the first population-based optimization method that has been adopted for the FS problem and was applied in many researches [27], [28]. GA finds an optimal solution by applying evolutionary operators such as crossover, mutation, and selection to the population. The authors in [28] firstly ranked features according to a filter criterion and then applied GA on the high-rank features to reduce the search space. Moslehi and Haeri [29] used the GA along with the PSO to achieve better performance. After integrating the populations obtained from these algorithms, the best solutions from the integrated population were selected. In [30], a bi-objective GA is used for an ensemble-based feature selection technique. The boundary region analysis and the multivariate mutual information were considered objective functions to select informative features.

Another population-based optimization method that has been widely used for solving FS problems in the literature is PSO. It was developed by Eberhart and Kennedy [31] for dealing with search and optimization problems. To increase the search capability of selecting distinctive features, a hybrid PSO-based FS algorithm with a local search strategy (called HPSO-LS) was proposed in [32]. The local search strategy provided by employing the correlation information of the features helps the search process to select less the best features. Amoozegar and Minaei-Bidgoli [13] proposed a multi-objective FS algorithm, namely RFPSOFS, that ranked the features based on their occurrences in the archive set. Then, the archive set was refined according to these rated features. Additionally, these ranks were also involved in updating the particle position vector and caused the particles to move purposefully. Most PSO-based FS methods used fixed-length representations that caused high computational costs for high dimensional data. The first variable and dynamic length representation for the PSO-based FS method (called VLPSO) was proposed in [9]. It enables the particles to have different lengths. In this way, the swarm was divided into several divisions so that each division had maximum length and the length of divisions are different from each other. Besides, the features were sorted in descending order of relevance. Then, a division by shorter length considers the top rank features for the selection process. To avoid getting stuck in the local optimum, the length of each division can be changed during the evolution process. This algorithm improved the performance of PSO by concentrating its search on reduced space and more fruitful areas.

ACO is one of the most well-known population-based optimization approaches proposed in 1999 [33] and has shown promise in the FS problem. For example, it was used with ANN in [18] for application in text FS that contains high dimensional data. In this method, two global and local rules were presented for updating the pheromone level. The global updating rule helps the algorithm to generate feature subsets with a low rate of classification error. Moreover, the local updating rule gives a chance for unrelated features which have not been investigated formerly to be selected and thus prevents early convergence. Generally, in ACO-based FS methods, there are multiple paths for a specific subset which causes an uneven distribution of pheromones sediments. To overcome this problem, Ghosh et al. [34] assigned pheromones sediments to nodes instead of edges between nodes. They proposed a wrapper-filter FS (WFACOFS) method to reduce the computational complexity. The algorithm generates the feature subsets using a filter method and then evaluates them by a classifier. Additionally, a fitness-based memory has been presented to keep the best solutions. So, in this way, FS is performed in a multi-objective manner.

In comparison with mentioned optimization algorithms, ABC was proposed later by Kraboga [35]. It deals with the optimization tasks using a vector-based representation which is appropriate for solving the FS problems. The authors in [36] have been applied GA crossover and mutation to their multi-objective ABC-based method that was incorporated with the non-dominated sorting process. Moreover, they utilized both binary and continuous representations. Kuo et al. [20] proposed an ABC-based that selects relevant features and simultaneously optimizes the SVM parameters. A cost-sensitive ABC-based feature selection approach called TMABC-FS was proposed in [21]. In multi-objective modeling of the problem, the feature cost is minimized, and the classification accuracy is maximized. The main contributions of the paper are to introduce two new operators, namely *diversity-guiding* and *convergence-guiding* searches for the onlooker and employed bees, respectively. Furthermore, it considers two archive sets, including *leader* and *external* archives, to improve the search procedure of different kinds of bees.

Estimation of distribution algorithm (EDA) is another population-based optimization method used in solving the FS problems. In [37], the EDA was applied for multi-objective feature selection in an intrusion detection system (IDS). The authors claimed that the proposed approach (MOEDAFS) has lower complexity and higher classification accuracy. The compact genetic algorithm (cGA) [38] is an EDA-based method that represents the population in keeping with the estimated probabilistic model over the set of solutions instead of genetic operators (crossover and mutation) in the traditional genetic algorithm. This method has also been applied for solving the FS problem in [39].

Other population-based optimization methods also have been used for solving the FS problem. The Cuckoo search algorithm in [23], firefly optimization in [40], and bat algorithm in [41] are some recent studies that tried to improve the performance of the FS problem. To summarize this section, Table 1 compares the specifications of the surveyed methods.

**Table 1. Comparison of the state-of-the-art FS methods that utilized population-based optimization.**

| Method | Population-based algorithm | Type | Representation method | The number of objectives | Classifier |
|---|---|---|---|---|---|
| HGA-NN [28] | GA | Wrapper | Binary | SO | ANN |
| HGP-FS [29] | GA, PSO | Hybrid | Continues | SO | ANN |
| Ensemble-FSGA [30] | GA | Filter | Binary | MO | - |
| HPSO-LS [32] | PSO | Hybrid | Continues | SO | KNN |
| RFPSOFS[13] | PSO | Wrapper | Continues | MO | KNN |
| VLPSO [9] | PSO | Hybrid | Continues | SO | KNN |
| ACO-ANN [18] | ACO | Wrapper | Continues | SO | ANN |
| WFACOFS [34] | ACO | Hybrid | Continues | SO | KNN/ANN |
| Hancer et al. [36] | ABC | Wrapper | Binary/ Continues | MO | KNN |
| ABC-SVM-DT [20] | ABC | Wrapper | Continues | SO | SVM/DT |
| TMABC-FS [21] | ABC | Wrapper | Continues | MO | KNN |
| MOEDAFS [37] | EDA | Wrapper | Binary | MO | - |
| cGA-FS [39] | cGA | Wrapper | Continues | SO | Naive Bayes |

Despite all advantages, the population-based optimization methods for solving FS problem suffer from several limitations that can be discussed in two directions:

1) High time complexity: large search space and consequently the large number of fitness function evaluations can lead to this problem.

2) Interaction between features: a feature may be weakly relevant to the target class individually, but the classification performance can be improved using some complementary features. Moreover, a feature may be relevant by itself, but it caused decreased classification performance when used with some other features.

To tackle the mentioned challenges of feature selection methods that utilized population-based optimization, the main contribution of this paper is to propose a correlation-aware EDA-based method and apply it for solving the FS problem. In the next section, the structure of the proposed method will be described in detail.

## 3. The proposed method

In this paper, we present a correlation-aware feature selection algorithm that not only considers the importance of each feature alone, but also can deal with the interaction between features. A good FS method should select a subset that features have minimum correlation and at the same time increases the classification performance. Therefore, the proposed method has the capability of considering the complementary features, and consequently, the classification performance will be improved. In addition to this, as we know, one of the main limitations of the wrapper-based FS methods, which utilized population-based optimization approaches is suffering from high complexity of time due to a large number of fitness function evaluations. Fortunately, the proposed method generates only two individuals in each iteration. Like the cGA, the generated individuals compete in each iteration of the algorithm based on a fitness function to determine the *winner* and the *loser*. These two individuals evolve during the algorithm to find the best solution. The number of features for each individual is determined based on our guiding technique, which will be discussed in section 3.5. In this technique, the number of features for each individual is determined randomly by chi-square distribution with $d$ degrees of freedom in each iteration, where $d$ is the number of *winner*'s features. In this way, the best value for $d$ is determined by evolution process, too.

To consider both the effects of each feature alone and the interaction between features in the proposed method, we define two data structures. The first one is the *significance vector (SV)* with size $n$, and the second one is the *interaction matrix (IM)* with size $n \times n$, where $n$ is the number of features. The $SV(i)$ represents the goodness of the corresponding feature $i$, while the $IM(i, j)$ denotes the goodness of simultaneous presence of two features $i$ and $j$ in the final solution. All elements of *SV* and *IM* are initialized by one to give an equal chance of selection to the features. Then, two individuals are generated using the conditional probabilities, which is one of the main advantages of the proposed method. The generated individuals then compete, and the *winner* and the *loser* are determined. Finally, *SV* and *IM* are updated using our update procedure that will be described in the following this section. The pseudocode of the proposed method is described in Figure 1.

---

        /\***Step 1: Initialization**\*/

1      Initialize the vector *SV* and the matrix $IM_{n \times n}$, as:

$$SV(i) = 1; IM(i, j) = 1; \ i, j = 1, \ldots, n.$$

2      *The_best_subset* = [];

3      **for** $i = 1$ to *iter*

        /\***Step 2: Generating two individuals *a* and *b***\*/

| | |
|---|---|
| | Select the first feature for each of *a* and *b* (denoted by *X*) with roulette wheel using probabilities *P*: |
| 4 | $$P(X_j^1) = SV(j) / \sum_{k=1}^{n} SV(k);$$ |
| 5 | Select the number of features for each individual using chi-square distribution: $$s \sim \chi^2(d);$$ |
| 6 | **for** *k* =2 to *s* |
| 7 | Select k-th feature for each of *a* and *b* (denoted by *X*) with roulette wheel using probabilities *P*: $$P(X_j^k \mid a_l \in A) = \frac{\left( \prod_{X_l \in A} IM(X_j, X_l) \right) \times SV(X_j)}{\left( \sum_{X_z \in A} \prod_{X_l \in A} IM(X_z, X_l) \times SV(X_z) \right)};$$ |
| 8 | **End for** |
| | **/*Step 3: Competition*/** |
| 9 | [*winner, loser*] = compete (*a,b*); |
| 10 | **if** (fitness(*winner*) > fitness (*The_best_subset*)) |
| 11 | *The_best_subset = winner;* |
| 12 | **End if** |
| | **/*Step 4: Update *SV* and *IM* using the update procedure*/** |
| 13 | **/*Step 5: Update the number of features *s* and *t* by our guiding technique*/** |
| 14 | **End for** |
| 15 | **Return** *The_best_subset* as the final solution; |

**Fig. 1. Pseudocode of the proposed method.**

## 3.1. Initialization

In the first step, all the elements of the significance vector *SV* and the interaction matrix *IM* are set to 1. It should be noted that the *SV* and *IM* data structures are used to determine the probability of selecting the features in the following steps. By assigning equal values to them, all features have the same chance to be selected at the beginning of the algorithm.

## 3.2. Generating two individuals

To select the best features subset, the algorithm generates two independent individuals in each iteration. The probability of selecting the first feature for each individual *a* and *b* is determined by its associated significance value divide by the sum of the significant value of all features:

$$P(X_j^1) = \frac{SV(j)}{\sum_{k=1}^{n} SV(k)}, \forall \ j = 1, ..., n. \tag{1}$$

It should be noted that these calculations are performed separately for each individual *a* and *b*, which are denoted by variable *X*. Then, the first feature is selected using the roulette wheel mechanism for each individual based on the calculated probabilities in Equation (1). Greater probability value of $P(X_j^1)$ gives more chance to *j*-th feature to be selected as the first one. In the first iteration, the probability of selecting each feature is $1/n$. However, since the *SV* is updated at the end of each iteration, it will be different for each feature in the successive iterations.

Similarly, the *k*-th feature for each individual is selected using the roulette wheel mechanism based on the following conditional probabilities:

$$P(X_j^k \mid X_l \in A) = \frac{\left( \prod_{X_l \in A} IM(X_j, X_l) \right) \times SV(X_j)}{\left( \sum_{X_z \in A} \sum_{X_l \in A} \prod IM(X_z, X_l) \times SV(X_z) \right)}, \quad \forall k = 2, ..., s, \tag{2}$$

where in Equation (2), $P(X_j^k \mid X_l \in A)$ denotes the probability of the *j*-th element of each individual $(X_j)$ to be selected as the *k*-th feature, when a set of features $A$ is selected in the previous iterations. The numerator represents the goodness of simultaneous presence of $X_j$ and previously selected features and the significance value of $X_j$. The denominator represents the goodness of simultaneous presence of unselected features $\bar{A}$ and previously selected features $A$ and the significance value of each unselected feature. This process continues until *s* features are selected for each individual. It should be noted that the variable *s* is calculated for each *a* and *b*, separately. We will discuss determining the variable *s* in subsection 3.5. It should be noted that however the *IM* reflects the interaction between only two features, the introduced probability in Equation (2) considers the goodness of selecting one feature given selecting a subset of selected features.

### 3.3 Competition

The generated individuals *a* and *b* from the previous step are then evaluated according to the following fitness function:

$$fitness = \frac{accuracy}{SFR},$$

$$\tag{3}$$

$$SFR = \frac{number\ of\ selected\ features}{total\ number\ of\ features}.$$

According to Equation (3), the fitness value of each individual is calculated by the classification accuracy achieved by the corresponding selected features of the individual divide by the selected feature rate (SFR). In this way, we can deal with both increasing the classification performance and decreasing the number of selected features. To calculate the accuracy of each candidate solution, we benefit from the advantages of support vector machines as the classifiers, including the generalization ability, strong theoretical foundations, absence of local minima, and robustness against noise. After evaluation, the individual with greater (smaller) fitness is called the *winner* (*loser*). The *winner* and the *loser* are binary vectors with length *n*. Thus, $w_i=1$ indicates that the *i*-th feature has been selected by the *winner*, while $w_i=0$ means that the *i*-th feature has not been selected. Similarly, elements of the *loser* vector are considered as $l_i$, which can be either 0 or 1. They are used to update the *SV* and *IM* in the next step of the proposed algorithm. If the current winner's fitness is greater than the fitness of the best solution found so far, the best solution is replaced by the current winner.

### 3.4 Update procedure

In this step, *SV* and *IM* should be updated using the obtained *winner* and *loser*. We update these two data structures using Table 2 and Table 3, respectively.

**Table 2. The update procedure of *SV* based on the *winner* and the *loser* vectors.**

| loser / winner | $l_i = \mathbf{0}$ | $l_i = \mathbf{1}$ |
|---|---|---|
| $w_i = \mathbf{0}$ | $SV$ | $SV-$ |
| $w_i = \mathbf{1}$ | $SV+$ | $SV$ |

As shown in Table 2, each element of the *SV* is updated based on the corresponding values of the *winner* and the *loser* vectors. In the case of both the *winner* and the *loser* have selected a feature or both have not selected it, we cannot decide whether selecting or not selecting the corresponding feature is good or not. Therefore, the corresponding value in *SV* remains unchanged. Meanwhile, if the *loser* selects a feature while the *winner* does not select it, we decrease the chance of selecting it by a predefined value between zero and one. We name this value as the *change factor*. It is used for strengthening or weakening the significance value of feature *i*. Finally, when the *winner* selects a feature while the *loser* does not select it, the chance of its selection is increased by the *change factor*.

**Table 3. The update procedure of *IM* based on the *winner* and the *loser* vectors.**

| loser / winner | $l_i, l_j = \mathbf{0,0}$ | $l_i, l_j = \mathbf{0,1}$ | $l_i, l_j = \mathbf{1,0}$ | $l_i, l_j = \mathbf{1,1}$ |
|---|---|---|---|---|
| $w_i, w_j = \mathbf{0,0}$ | $IM$ | $IM$ | $IM$ | $IM-$ |
| $w_i, w_j = \mathbf{0,1}$ | $IM$ | $IM$ | $IM$ | $IM--$ |
| $w_i, w_j = \mathbf{1,0}$ | $IM$ | $IM$ | $IM$ | $IM--$ |
| $w_i, w_j = \mathbf{1,1}$ | $IM+$ | $IM++$ | $IM++$ | $IM$ |

In Table 3, the selecting status of each pair of features $i$ and $j$ is compared, and some updates on *IM* are performed based on differences between *winner* and *loser*. In the following, we discuss how to update the *IM* in some of the states. The update procedure of the remaining states is similar.

- *$w_i=l_i$ and $w_j=l_j$*: for all four states with this condition, no updates on *IM* are made because the *winner* and the *loser* did the same for selecting the features $i$ and $j$. Therefore, we cannot decide whether selecting or not selecting the features $i$ and $j$, together, is good or not.

- $(w_i, w_j=0,0)$ and $(l_i, l_j=0,1)$: in this state, both the *winner* and the *loser* have not selected the feature $i$. Thus, we cannot decide about the goodness of selecting this feature. However, since the *loser* has selected the $j$-th feature and the *winner* has not selected it, we decrease the chance of selecting it by the introduced *change factor*.

- $(w_i, w_j=0,0)$ and $(l_i, l_j=1,1)$: here, the *winner* has not selected any of $i$ and $j$, while the *loser* has selected both of them. Therefore, we decrease the $IM(i, j)$ by the *change factor*.

- $(w_i, w_j=0,1)$ and $(l_i, l_j=0,0)$: in this state, the *winner* has selected the $j$-th feature, and none of the features $i$ and $j$ have been selected by the *loser*. Hence, the chance of simultaneous appearing the features $i$ and $j$ remains unchanged.

- $(w_i, w_j=0,1)$ and $(l_i, l_j=1,1)$: since the *winner* has not selected the feature $i$ and $j$ together and, at the same time, the *loser* has selected both of them, and we decrease the $IM(i, j)$ by a stronger value than the *change factor* (i.e., two or three times higher), because in this state, we are more confident in decreasing or increasing the chance of selecting the features.

Finally, it should be emphasized that we run the update procedure only when that the new *winner* has better fitness value than the previously generated *winner* to avoid convergence to bad solutions.

### 3.5. Update the number of features for individuals by our guiding technique

One of the advantages of the proposed algorithm is that the number of features for each of two individuals in each iteration is determined based on the number of *winner*'s features. In other words, we guide the algorithm to select optimum features. To this aim, the number of each individual, which is denoted by $s$, in Figure 1, are random numbers determined by chi-square distribution with $d$ degrees of freedom, where $d$ is the number of *winner*'s features. For the first iteration, $d$ is initialized to $n/2$. The expected value for the chi-square is equal to $d$ but it is possible to take the values for the number of features $s$ more or less as well. Since the variable $d$ is defined as the number of *winner*'s features, which will be updated in each iteration, the number of selected features of the final solution will be optimized during the evolution process. Moreover, this guiding mechanism ensures that the number of features for each individual does not exceed the value determined by the chi-square distribution. This can directly increase the convergence speed of the algorithm due to the limitation on the number of features for each individual. Additionally, it can help the proposed algorithm to select fewer features. In Section 4.4, we will discuss the results of applying this technique to real-world datasets.

## 4. Experimental results

In this section, we evaluate our method using different datasets and compare it with state-of-the-art studies. It should be noted that the proposed algorithm has been implemented using MATLAB® 2018a. Besides, all the experiments are performed on a machine with 2.60 GHz Intel Core i7 processor and 6.0GB of DDR3 memory. We will compare our proposed method with GA-SVM, cGA-FS [39], WFACOFS [34], MOEDAFS [37], and RSVM-SBS [42]. To select the best feature subset, GA-SVM utilizes the genetic algorithm as optimization technique and support vector machines as the fitness function. The following three approaches were described in Table 1 in Section 2. Moreover, RSVM-SBS combines the sequential backward search (SBS) with noise-aware support vector machines, namely RSVM, to deals with the FS problem in the presence of outliers. In all of the experiments, the datasets are firstly divided randomly into 75% training, and 25% testing sets, and then they are used for all of the methods for the comparisons. Finally, the *change factor* value in our updating procedure is set to 0.01.

### 4.1 Datasets

The details of datasets used to assess the proposed approach and its comparisons to other approaches in the literature are summarized in Table 4. All of the datasets in our experiments are obtained from the UCI Repository [43]. These datasets are from various fields and can be categorized based on the number of features into three groups: small, medium, and large. A dataset with less than ten features is considered small, while it is placed in the large category if its number of features is more than 100, and otherwise, it will be a medium dataset. We will test our algorithm on three small, seven medium, and four large datasets.

**Table 4. Details of datasets.**

| Dataset | Size | Number of samples | Number of features | Number of classes |
|---|---|---|---|---|
| Breast Cancer | Small | 699 | 9 | 2 |
| Glass | Small | 214 | 9 | 6 |
| Heart | Medium | 270 | 13 | 2 |
| Wine | Medium | 178 | 13 | 3 |
| Segmentation | Medium | 2310 | 19 | 7 |
| German | Medium | 1000 | 24 | 2 |
| Ionosphere | Medium | 351 | 34 | 2 |
| Soybean-small | Medium | 47 | 35 | 4 |
| Sonar | Medium | 208 | 60 | 2 |
| Hill-valley | Large | 1212 | 100 | 2 |
| Musk1 | Large | 476 | 167 | 2 |
| Arrhythmia | Large | 452 | 279 | 16 |
| Isolet5 | Large | 1559 | 617 | 26 |

### 4.2 Performance metrics

To measure the performance of different methods, some well-known metrics are used, including accuracy, precision, recall, F1-score, and a metric which is introduced in [42], i.e., the product of accuracy (*ACC*) rate and the percentage of discarded features (*PDF*). These metrics are defined as follows:

$$Accuracy = \frac{TP + TN}{TP + TN + FP + FN},\tag{4}$$

$$Precision = \frac{TP}{TP \times FP},\tag{5}$$

$$recall = \frac{TP}{TP + FN},\tag{6}$$

$$F1-score = \frac{2TP}{2TP + FP + FN},\tag{7}$$

$$ACC \times PDF = Accuracy \times percentage\ of\ discarded\ features,\tag{8}$$

where TP, TN, FP, and FN are the number of true positives, true negatives, false positives, and false negatives, respectively.

### 4.3 The results on a synthetic dataset with correlated features

One of the main contributions of the proposed method is that it can deal with the correlation between features. To investigate this, we generate a synthetic dataset. This dataset consisted of 10 correlated features and 250 samples with random values in two classes. There is no correlation between the first six features, but the values of other features are generated as follows:

$$f_7 = 10 \times f_1, \quad f_8 = (f_2 + 3 \times f_3), \quad f_9 = f_4, \quad f_{10} = f_5 \,/\, 1000,\tag{9}$$

where $f_i$ denotes the $i$-th feature. A good FS method should not select the correlated features. For example, based on Equation (9), only one of $f_1$ or $f_7$ should appear in the final solution for our synthetic dataset. After running the proposed method on the synthetic dataset, it selects $\{f_1, f_2, f_3, f_4, f_{10}\}$ as the best feature set with accuracy 98.39%. The proposed method has not selected any correlated feature. Table 5 summarizes the obtained results for different methods on the synthetic dataset.

**Table 5. The accuracy (%) for different methods on the synthetic dataset.**

| Method | Accuracy | Selected features |
|---|---|---|
| GA-SVM | 98.38 | $\{f_1, f_5, f_6, f_7, f_8, f_9\}$ |
| cGA-FS | 80.76 | $\{f_1, f_2, f_3, f_4, f_8, f_9\}$ |
| WFACOFS | 79.03 | $\{f_1, f_3, f_5, f_6, f_7, f_9, f_{10}\}$ |
| MOEDAFS | 83.87 | $\{f_1, f_4, f_5, f_7, f_8, f_9, f_{10}\}$ |
| RSVM-SBS | 81.69 | $\{f_1, f_2, f_5, f_6, f_7, f_8, f_{10}\}$ |
| The proposed Method | 98.39 | $\{f_1, f_2, f_3, f_4, f_{10}\}$ |

As shown in Table 5, the proposed method has selected the least number of features. Meanwhile, it has reported higher accuracy, and most importantly, has not selected any correlated feature. However, the other methods failed to do so. For example, cGA-FS has selected $f_4$ and $f_9$, simultaneously. Similarly, WFACOFS has selected $f_1$ and $f_7$ as well as $f_5$ and $f_{10}$. The other methods also have behaved similarly. To better understand how the proposed approach obtains these results, we should review the role of the probability scheme, which is introduced in the previous section. Figure 2 represents the heatmap matrix (HM) of our introduced conditional probabilities determined by Equation (2) after 1000 runs of the proposed method on the synthetic dataset. The correlate features are highlighted in bold.

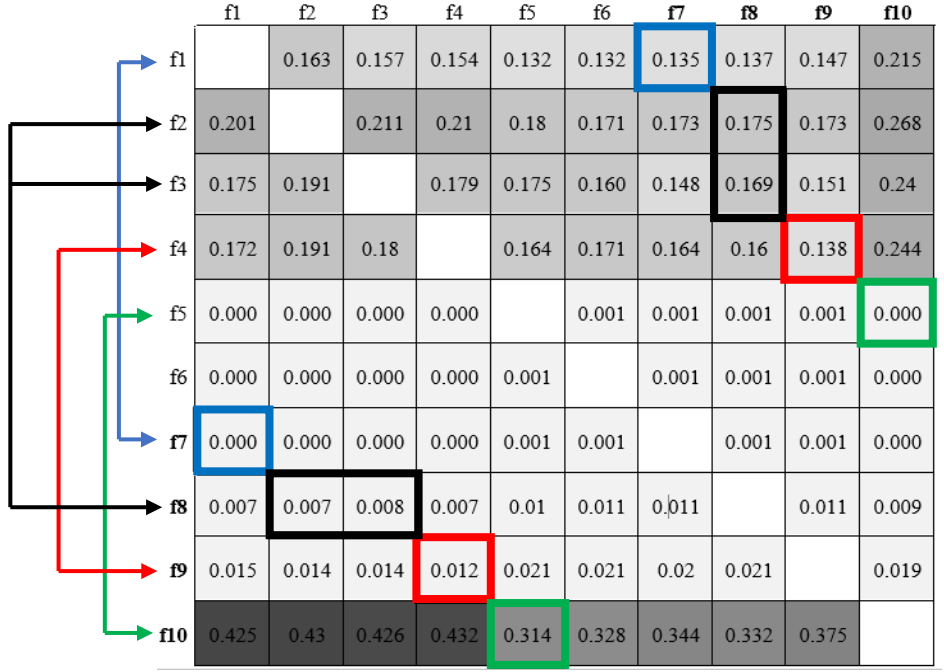

|      | f1    | f2    | f3    | f4    | f5    | f6    | f7    | f8    | f9    | f10   |
|------|-------|-------|-------|-------|-------|-------|-------|-------|-------|-------|
| f1   |       | 0.163 | 0.157 | 0.154 | 0.132 | 0.132 | 0.135 | 0.137 | 0.147 | 0.215 |
| f2   | 0.201 |       | 0.211 | 0.21  | 0.18  | 0.171 | 0.173 | 0.175 | 0.173 | 0.268 |
| f3   | 0.175 | 0.191 |       | 0.179 | 0.175 | 0.160 | 0.148 | 0.169 | 0.151 | 0.24  |
| f4   | 0.172 | 0.191 | 0.18  |       | 0.164 | 0.171 | 0.164 | 0.16  | 0.138 | 0.244 |
| f5   | 0.000 | 0.000 | 0.000 | 0.000 |       | 0.001 | 0.001 | 0.001 | 0.001 | 0.000 |
| f6   | 0.000 | 0.000 | 0.000 | 0.000 | 0.001 |       | 0.001 | 0.001 | 0.001 | 0.000 |
| f7   | 0.000 | 0.000 | 0.000 | 0.000 | 0.001 | 0.001 |       | 0.001 | 0.001 | 0.000 |
| f8   | 0.007 | 0.007 | 0.008 | 0.007 | 0.01  | 0.011 | 0.011 |       | 0.011 | 0.009 |
| f9   | 0.015 | 0.014 | 0.014 | 0.012 | 0.021 | 0.021 | 0.02  | 0.021 |       | 0.019 |
| f10  | 0.425 | 0.43  | 0.426 | 0.432 | 0.314 | 0.328 | 0.344 | 0.332 | 0.375 |       |

**Fig. 2. The conditional probability heatmap matrix. The conditional probability heatmap matrix. The correlated features have been shown with different colors. Smaller values in each row indicate more correlated between corresponding features.**

Each element $HM(i,j)$ in Figure 2 denotes the probability of selecting the $j$-th feature given the $i$-th feature. The sum of the conditional probabilities at any column is equal to one, as expected. As we can conclude from the heatmap matrix, the probabilities of selecting the correlated features have decreased using our update procedure (see Section 3.4). For example, $HM(5,10)$ and $HM(10,5)$ have the lowest values in their rows. It means that when we select $f_5$, the probability of selecting $f_{10}$ is less than the other remaining features. Likewise, when $f_{10}$ is selected by the proposed algorithm, the probability of selecting $f_5$ is less than the other features. Smaller values in each row indicate more correlation between corresponding features. As we described earlier, however the *IM* reflects the interaction between only two features, the proposed algorithm considers other interactions (see Section 3.2). For example, as shown in the first row of *HM* in Figure 2, the value of $HM(1,6)$ and $HM(1,5)$ are lower than $HM(1,7)$ but the algorithm has not selected $f_1$ and $f_7$, simultaneously.

### 4.4 The results on real-world datasets

Here, we compare our proposed method with other approaches on real-world datasets. Table 6 to Table 10 summarize the obtained results in terms of different performance metrics after ten independent runs. In each run of the population-based optimization FS methods and our proposed method, the results are reported after 500 fitness function evaluations. The best results in each table are highlighted in boldface.

**Table 6. The average accuracy (%) metric for different methods on different datasets.**

|  | All-features | GA-SVM | cGA-FS | WFACOFS | MOEDAFS | RSVM-SBS | The proposed Method |
|---|---|---|---|---|---|---|---|
| Breast Cancer | 70.29 | 96.57 | 94.85 | 97.71 | 96.00 | 96.58 | **98.85** |
| Glass | 56.60 | 62.26 | 67.92 | 69.81 | 69.81 | **72.56** | 71.70 |
| Heart | 82.09 | 88.06 | 85.07 | 83.58 | 89.55 | 83.75 | **91.04** |
| Wine | 93.18 | 97.89 | 97.22 | 97.72 | 98.83 | 98.04 | **99.36** |
| Segmentation | 93.07 | 93.58 | 87.69 | **95.32** | 94.28 | 91.76 | 94.80 |
| German | 73.60 | 82.40 | 79.20 | 77.60 | 80.40 | 78.91 | **82.80** |
| Ionosphere | 86.36 | 92.05 | 88.63 | 94.31 | 89.77 | 90.69 | **95.45** |
| Soybean-small | 100.00 | 100.00 | 90.00 | 100.00 | 100.00 | 100.00 | **100.00** |
| Sonar | 73.08 | 78.84 | 82.69 | 90.38 | 90.38 | 89.03 | **92.30** |
| Hill-valley | 48.18 | **92.73** | 59.07 | 51.48 | 53.46 | 61.79 | 71.61 |
| Musk1 | 82.35 | **92.43** | 68.06 | 86.55 | 86.55 | 82.74 | 84.03 |
| Arrhythmia | 51.32 | **74.33** | 61.94 | 60.17 | 61.06 | 63.97 | 72.66 |
| Isolet5 | 76.40 | **95.38** | 90.76 | 82.05 | 84.10 | 88.45 | 90.00 |

As shown in Table 6, the proposed method reported the best accuracies in eight datasets. GA-SVM obtained better accuracies on four datasets in comparison to the proposed method. Finally, WFACOFS and RSVM-SBS also had better results in Segmentation and Glass datasets, respectively. We should emphasize that high accuracy does not necessarily indicate the efficiency of a feature selection method. For example, considering the hill-valley dataset, GA-SVM selects 92 features from all 100 features. However, the proposed method selects only 24 features. In the following this section, we will report the best number of selected features by each method. Table 7 summarizes the average precisions of different methods on different datasets.

**Table 7. The average precision (%) metric for different methods on different datasets.**

|  | All-features | GA-SVM | cGA-FS | WFACOFS | MOEDAFS | RSVM-SBS | The proposed Method |
|---|---|---|---|---|---|---|---|
| Breast Cancer | 71.53 | 96.67 | 94.56 | 97.81 | 95.98 | 96.80 | **98.80** |
| Glass | 43.53 | 63.30 | 54.19 | 61.42 | 59.59 | 66.73 | **68.75** |
| Heart | 81.91 | 86.22 | 84.85 | 83.88 | 88.51 | 86.46 | **91.70** |
| Wine | 93.89 | 97.87 | 97.43 | 98.14 | 98.43 | 97.52 | **99.86** |
| Segmentation | 93.28 | 93.35 | 88.03 | **95.40** | 94.46 | 73.19 | 94.81 |
| German | 76.26 | 72.89 | 70.80 | 70.29 | 69.23 | 71.37 | **77.24** |
| Ionosphere | 82.38 | 89.45 | 77.27 | 93.73 | 85.61 | 90.07 | **94.29** |
| Soybean-small | 100.00 | 100.00 | 88.12 | 100.00 | 100.00 | 100.00 | **100.00** |
| Sonar | 76.67 | 80.00 | 82.68 | 90.29 | 90.74 | 91.61 | **92.15** |
| Hill-valley | 48.13 | **92.79** | 56.25 | 52.72 | 54.43 | 63.71 | 70.43 |
| Musk1 | 82.44 | **92.10** | 68.43 | 86.50 | 87.00 | 87.11 | 83.75 |
| Arrhythmia | 29.87 | **55.14** | 38.76 | 13.61 | 18.18 | 33.08 | 36.20 |
| Isolet5 | 67.96 | **95.82** | 89.73 | 81.75 | 83.83 | 80.57 | 90.07 |

As reported in Table 7, our proposed approach outperformed the other methods in eight datasets. However, GA-SVM showed the better results on four datasets in comparison to the proposed method. Moreover, WFACOFS reported better results on Segmentation. Table 8 reports the average recall of different methods on different datasets.

**Table 8. The average recall (%) metric for different methods on different datasets.**

| | All-features | GA-SVM | cGA-FS | WFACOFS | MOEDAFS | RSVM-SBS | The proposed Method |
|---|---|---|---|---|---|---|---|
| Breast Cancer | 68.88 | 96.31 | 94.24 | 96.80 | 94.38 | 95.35 | **98.80** |
| Glass | 33.85 | **70.58** | 63.31 | 67.60 | 63.40 | 70.35 | 59.58 |
| Heart | 84.10 | 89.95 | 83.88 | 84.18 | 90.05 | 89.06 | **90.95** |
| Wine | 92.90 | 96.94 | 98.61 | 96.96 | 97.66 | 98.68 | **98.81** |
| Segmentation | 93.93 | 93.43 | 88.45 | 95.76 | 94.89 | 81.50 | **95.90** |
| German | 70.01 | 75.50 | 81.25 | 75.40 | **82.77** | 75.96 | 80.30 |
| Ionosphere | 85.42 | 93.24 | 93.42 | 94.24 | 86.74 | 94.85 | **96.49** |
| Soybean-small | 100.00 | 100.00 | 86.22 | 100.00 | 100.00 | 100.00 | **100.00** |
| Sonar | 80.56 | 78.44 | 82.44 | 90.47 | 91.67 | 89.91 | **92.65** |
| Hill-valley | 47.91 | **92.81** | 60.59 | 53.21 | 54.65 | 66.76 | 80.26 |
| Musk1 | 82.33 | **92.35** | 68.88 | 86.50 | 87.32 | 90.31 | 83.58 |
| Arrhythmia | 24.12 | **54.27** | 40.08 | 16.16 | 38.77 | 43.34 | 46.78 |
| Isolet5 | 71.81 | **95.48** | 89.72 | 83.36 | 85.58 | 89.10 | 89.90 |

As can be seen from Table 8, our proposed algorithm achieved the best recall on seven datasets. However, GA-SVM and MOEDAFS methods reported better results than the proposed algorithm on five and two datasets, respectively. The F1-score results of different methods are represented in Table 9.

**Table 9. The average F1-score (%) metric for different methods on different datasets.**

| | All-features | GA-SVM | cGA-FS | WFACOFS | MOEDAFS | RSVM-SBS | The proposed Method |
|---|---|---|---|---|---|---|---|
| Breast Cancer | 70.58 | 96.79 | 93.87 | 97.58 | 95.47 | 96.24 | **98.89** |
| Glass | 36.48 | 66.42 | 56.91 | 64.32 | 60.84 | **67.58** | 64.78 |
| Heart | 83.09 | 88.34 | 83.97 | 84.03 | 89.07 | 86.72 | **92.25** |
| Wine | 92.83 | 97.82 | 96.67 | 97.04 | 98.65 | 98.29 | **99.46** |
| Segmentation | 93.91 | 92.89 | 88.39 | 95.16 | 94.28 | 78.13 | **95.28** |
| German | 73.26 | 73.97 | 75.58 | 72.59 | 75.21 | 73.39 | **77.91** |
| Ionosphere | 83.13 | 90.43 | 84.15 | 93.21 | 86.73 | 93.05 | **95.72** |
| Soybean-small | 100.00 | 100.00 | 87.08 | 100.00 | 100.00 | 100.00 | **100.00** |
| Sonar | 79.28 | 79.93 | 83.28 | 91.10 | 91.92 | 91.47 | **93.12** |
| Hill-valley | 47.37 | **93.02** | 58.71 | 53.33 | 54.91 | 65.57 | 75.39 |
| Musk1 | 81.85 | **91.69** | 68.12 | 85.97 | 86.79 | 88.15 | 83.14 |
| Arrhythmia | 25.98 | **54.31** | 38.70 | 14.07 | 24.04 | 36.81 | 40.10 |
| Isolet5 | 70.21 | **96.02** | 90.10 | 82.92 | 85.07 | 84.99 | 90.36 |

Table 9 results that our proposed approach obtained the best results on eight datasets. However, GA-SVM reported better results in terms of F1-score on four datasets. Furthermore, on Glass dataset, RSVM-SBS showed better results.

For the next experiment, we investigate the number of features selected by each method on different datasets. Table 10 compares the average and the best number of selected features for the final solution of different methods after ten runs.

**Table 10. The average and the best (in parentheses) number of selected features for different methods on different datasets.**

|  | All-features | GA-SVM | cGA-FS | WFACOFS | MOEDAFS | RSVM-SBS | The proposed Method |
|---|---|---|---|---|---|---|---|
| Breast Cancer | 9 (9) | 7.0 (5) | 6.9 (6) | 8.0 (7) | 7.3 (6) | 6.5 (6) | **4.3 (4)** |
| Glass | 9 (9) | 7.5 (7) | 8.0 (7) | 6.6 (4) | 6.5 (6) | 6.8 (6) | **6.5 (6)** |
| Heart | 13 (13) | 11.8 (11) | 7.0 (6) | 9.2 (8) | 6.6 (5) | 11.0 (10) | **4.6 (4)** |
| Wine | 13 (13) | 11.5 (10) | 8.0 (6) | 7.1 (6) | 9.7 (9) | 10.3 (9) | **4.3 (4)** |
| Segmentation | 19 (19) | 10.7 (9) | 10.7 **(7)** | 12.4 (9) | 8.9 (7) | 15.8 (14) | **8.7 (8)** |
| German | 24 (24) | 17.9 (12) | 17.1 (13) | 16.9 (13) | 18.2 (15) | 18.9 (16) | **14.8 (10)** |
| Ionosphere | 34 (34) | 26.3 (21) | 12.7 **(5)** | **10.8** (6) | 16.0 (6) | 18.5 (12) | 12.0 (8) |
| Soybean-small | 35 (35) | 17.2 (10) | 5.5 (4) | 16.8 (12) | 15.5 (14) | 16.3 (10) | **3.1 (2)** |
| Sonar | 60 (60) | 54.3 (44) | 42.0 (27) | 39.1 (33) | 19.3 (8) | 30.7 (17) | **13.7 (6)** |
| Hill-valley | 100 (100) | 68.1 (56) | 39.7 (26) | 56.9 (45) | 65.3 (46) | 63.4 (53) | **33.4 (24)** |
| Musk1 | 167 (167) | 143.2 (122) | 59.0 (47) | 42.5 (29) | 82.5 (71) | 56.0 (41) | **34.4 (22)** |
| Arrhythmia | 279 (279) | 188.9 (171) | 120.9 (95) | 44.7 (24) | 134.9 (115) | 87.2 (70) | **22.0 (15)** |
| Isolet5 | 617 (617) | 326.6 (273) | 297.2 (247) | 142.4 (97) | 392.8 (327) | 236.3 (181) | **141.3 (94)** |

As shown in Table 10, the average number of selected features achieved by the proposed algorithm is less than the other methods on all datasets except for Ionosphere. Additionally, the proposed algorithm outperforms the other methods in term of the best number of selected features except for Ionosphere dataset.

As discussed earlier, higher performance metrics which were summarized in Table 6 to Table 9, does not guarantee higher efficiency of a feature selection method. In addition to the reported performance metrics, a good FS method should optimize the number of selected features. The reason is that minimizing the number of selected features increases the generalizability of the model and decreases its complexity. Thus, we provide a criterion which is introduced in [42], namely the product of accuracy ($ACC$) rate and the percentage of discarded features ($PDF$). Table 11 summarizes the obtained results for different methods in term of this new performance metric.

**Table 11. The average ACC×PDF (%) metric for different methods on different datasets.**

|  | All-features | GA-SVM | cGA-FS | WFACOFS | MOEDAFS | RSVM-SBS | The proposed Method |
|---|---|---|---|---|---|---|---|
| Breast Cancer | 0.00 | 32.19 | 26.87 | 16.28 | 25.06 | 29.50 | **53.26** |
| Glass | 0.00 | 12.10 | 11.31 | **28.69** | 21.33 | 20.95 | 21.90 |
| Heart | 0.00 | 10.83 | 42.53 | 28.28 | 49.59 | 16.10 | **53.22** |
| Wine | 0.00 | 17.18 | 45.00 | 48.47 | 27.92 | 25.56 | **67.86** |
| Segmentation | 0.00 | 45.06 | 46.84 | 41.63 | 54.82 | 19.80 | **55.13** |
| German | 0.00 | 31.07 | 29.53 | 29.25 | 24.79 | 21.53 | **40.02** |
| Ionosphere | 0.00 | 28.01 | 65.55 | 66.00 | 60.72 | 50.01 | **67.37** |
| Soybean-small | 0.00 | 45.42 | 75.06 | 58.85 | 53.56 | 62.42 | **92.71** |
| Sonar | 0.00 | 14.25 | 35.13 | 36.07 | 70.56 | 53.63 | **77.14** |
| Hill-valley | 0.00 | 35.19 | 39.66 | 25.24 | 23.70 | 25.82 | **51.05** |
| Musk1 | 0.00 | 19.03 | 46.45 | 68.02 | 46.77 | 58.70 | **69.84** |
| Arrhythmia | 0.00 | 26.38 | 37.96 | 52.75 | 33.71 | 45.94 | **67.84** |
| Isolet5 | 0.00 | 49.03 | 50.73 | 66.13 | 35.03 | 58.53 | **72.83** |

The obtained results in Table 11 proved that the proposed approach increased the accuracy and simultaneously decreased the selected number of features. Our proposed approach achieved the best results on 12 out of 13 datasets.

To say more about the superiority of the proposed method, we rank different methods based on each performance metrics on all the datasets. Table 12 represents the count of achieved the best rank by each method.

**Table 12. The number of reported best results for different methods in terms of each performance metric on different datasets.**

|  | All-features | GA-SVM | cGA-FS | WFACOFS | MOEDAFS | RSVM-SBS | The proposed Method |
|---|---|---|---|---|---|---|---|
| Accuracy | 1 | 5 | 0 | 2 | 1 | 2 | **8** |
| Precision | 1 | 5 | 0 | 2 | 1 | 2 | **8** |
| Recall | 1 | 6 | 0 | 1 | 3 | 1 | **7** |
| F1-score | 1 | 5 | 0 | 1 | 2 | 2 | **8** |
| ACC×PDF | 0 | 0 | 0 | 1 | 1 | 0 | **12** |
| # of Selected Features | 0 | 0 | 0 | 1 | 2 | 0 | **13** |

From Table 12, it can be found that the proposed method has been reported the best rank more times than the other methods. For example, it achieved the best rank on eight datasets in terms of the accuracy metric. Additionally, for some metrics, the sum of the best ranks exceeds the total number of datasets, which means that more than one method achieves the best result on some datasets. It can be observed from Table 10 that the proposed method obtained promising results in terms of ACC×PDF. It proves that the proposed method has considered increasing the accuracy and simultaneously decreasing the number of selected features.

For the next experiment, Figure 3 represents the average ranks of different methods in terms of different performance metrics.

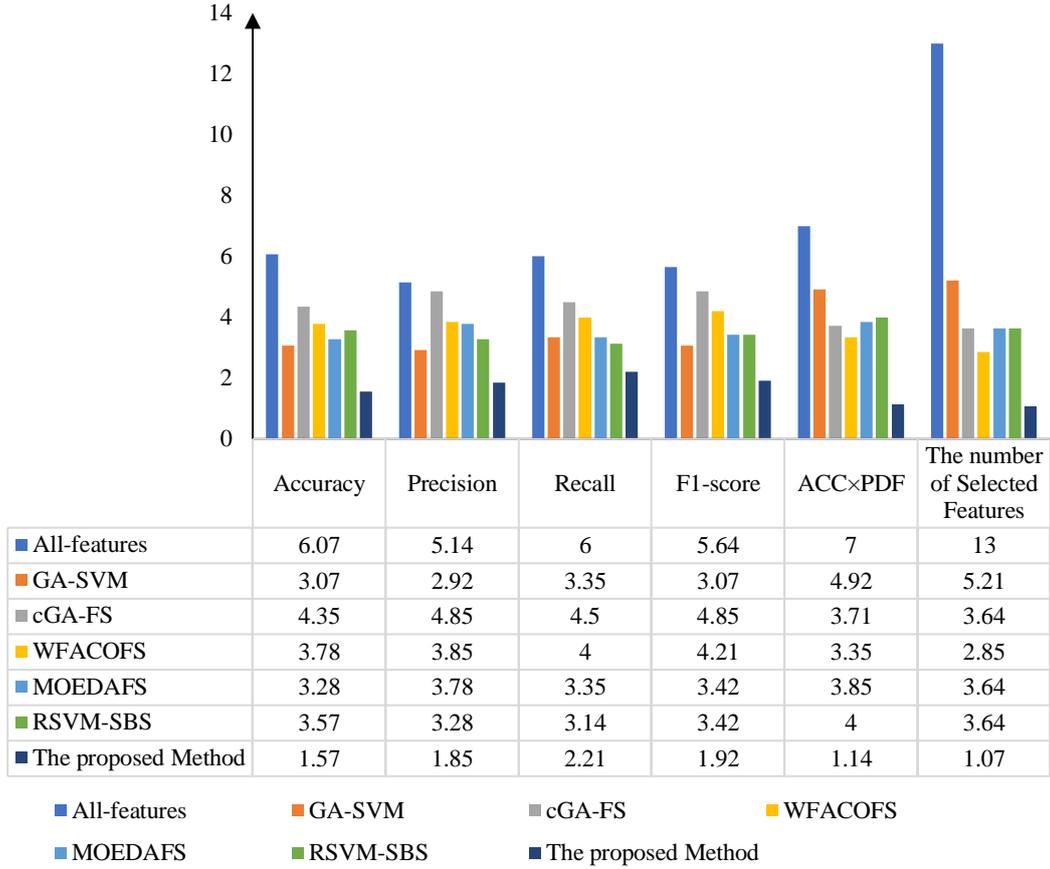

| | Accuracy | Precision | Recall | F1-score | ACC×PDF | The number of Selected Features |
|---|---|---|---|---|---|---|
| All-features | 6.07 | 5.14 | 6 | 5.64 | 7 | 13 |
| GA-SVM | 3.07 | 2.92 | 3.35 | 3.07 | 4.92 | 5.21 |
| cGA-FS | 4.35 | 4.85 | 4.5 | 4.85 | 3.71 | 3.64 |
| WFACOFS | 3.78 | 3.85 | 4 | 4.21 | 3.35 | 2.85 |
| MOEDAFS | 3.28 | 3.78 | 3.35 | 3.42 | 3.85 | 3.64 |
| RSVM-SBS | 3.57 | 3.28 | 3.14 | 3.42 | 4 | 3.64 |
| The proposed Method | 1.57 | 1.85 | 2.21 | 1.92 | 1.14 | 1.07 |

■ All-features ■ GA-SVM ■ cGA-FS ■ WFACOFS
■ MOEDAFS ■ RSVM-SBS ■ The proposed Method

**Fig. 3. The average ranks of different methods in terms of each performance metric on different datasets.**

From Figure 3, it can be seen that the proposed method has reported lower average ranks in all terms of performance metrics. It should be noted that the lower rank indicates better results.

To efficiency analysis of the experimental results obtained by different methods, two non-parametric statistical tests, namely Wilcoxon's signed-rank test [44] and Friedman's test [45], with a significance level of 0.05, are performed. The Wilcoxon's signed-rank test is utilized for pairwise performance evaluation between the proposed approach and the compared methods. Table 13 summarizes the Wilcoxon's signed-rank test results in terms of ACC×PDF metric.

**Table 13. P-value of Wilcoxon signed-rank test between the proposed approach and each other methods in terms of ACC×PDF.**

| Method | $p$-value |
|---|---|
| All-features | 0.0002 |
| GA-SVM | 0.0002 |
| cGA-FS | 0.0002 |
| WFACOFS | 0.0017 |
| MOEDAFS | 0.0002 |
| RSVM-SBS | 0.0002 |

Based on the obtained results from Table 13, the proposed approach shows a significant difference from the compared methods on all datasets.

The Friedman's test is also applied to evaluate the performance of all compared methods in terms of the ACC×PDF metric. After performing this test, the $p$-value was equal to 1.38E-9, which indicates that the overall performance of the proposed method is significantly better than the other methods.

### 4.4 Guiding technique analysis

As discussed in Section 3.5, our introduced guiding technique ensures that the number of features for each individual does not exceed the value determined by the chi-square distribution with $d$ degrees of freedom, where $d$ is the number of *winner*'s features. This can directly increase the convergence speed of the algorithm due to the limitation on the number of features for each individual. It can also help the proposed method to select fewer features. Figure 4 shows the average number of selected features for the *winner* after ten separate runs of the algorithm with 500 function evaluations. We perform this experiment on the Hill-valley dataset with 100 features. However, the other datasets also have similar trends and so we ignore reporting them.

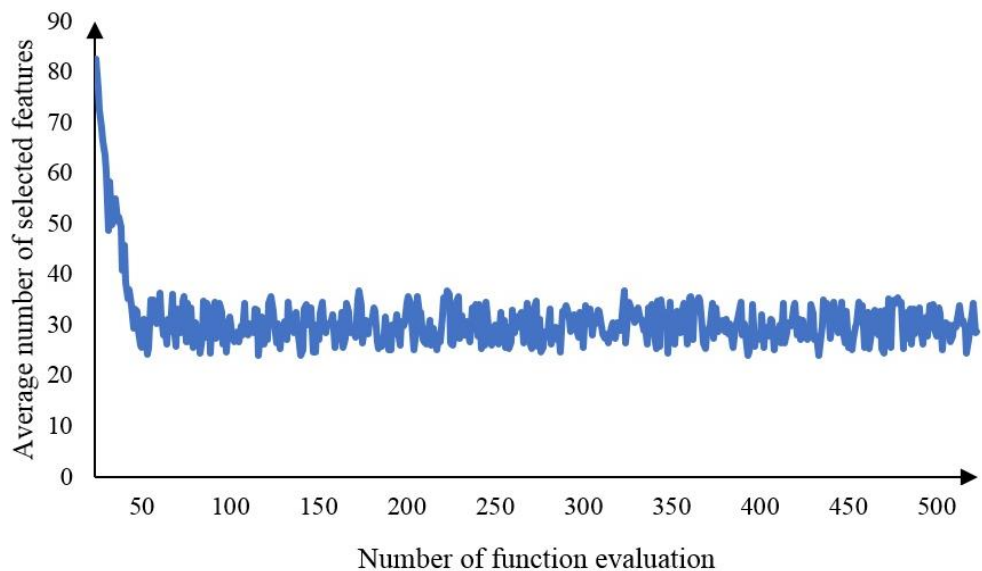

**Fig. 4. The average number of selected features of the *winner*, determined by the chi-square distribution for the Hill-valley dataset.**

As shown in Figure 4, the proposed method tends to select the minimum number of features almost at the beginning of the algorithm.

For the last experiment, we investigate the convergence speed of the proposed algorithm and compare it with other population-based optimization FS methods. Figure 5 shows the number of function evaluations to achieve the best average accuracies in the evolution process. The results are reported after ten separate runs of the algorithm with 500 function evaluations for the Heart dataset. It should be noted that the other datasets also have similar behavior and so we ignore reporting them.

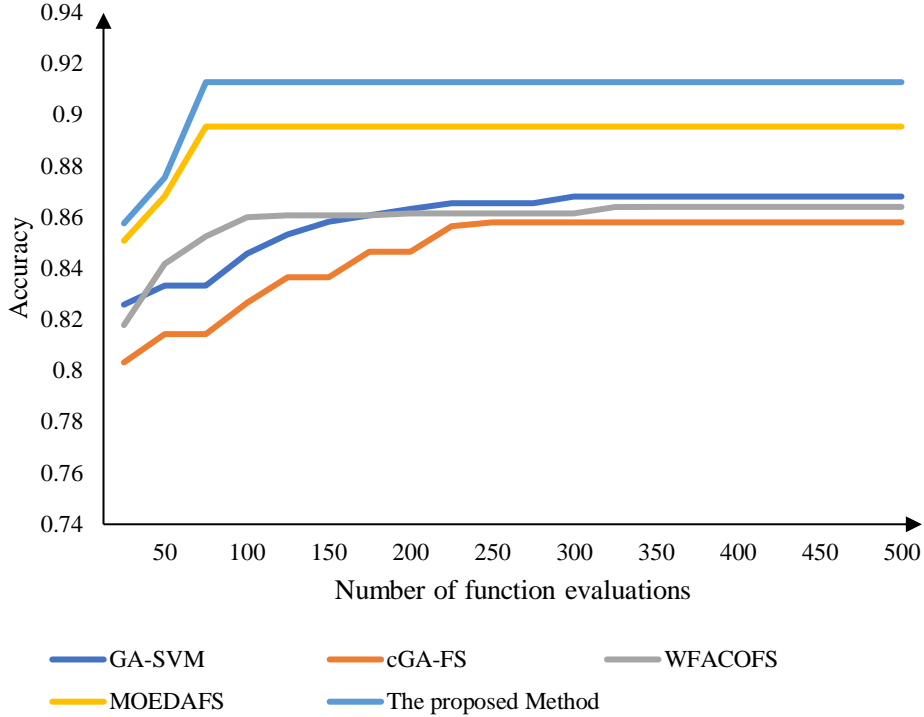

**Fig. 5.** The number of function evaluations to achieve the best average accuracies for the Heart dataset.

As shown in Figure 5, the proposed method has converged to a higher accuracy with fewer number of function evaluations. The proposed method reported the best answer in 73-th function evaluation, but GA-SVM, cGA-FS, WFACOFS, and MOEDAFS have achieved the best results with 300, 220, 350, and 75 function evaluations, respectively.

## 5. Conclusion and future works

Feature selection is one of the critical preprocessing steps in each machine learning application. It tries to find the optimal subset of informative features and consequently remove irrelevant ones. The main advantages of FS are reducing the time complexity and the cost of building the model, preventing over-fitting, increasing the generalizability of the trained classifier, and increasing its accuracy. However, FS suffers from a large search space and a correlation between features that severely affecting its performance. There are several studies in the literature for each of these challenges. Among them, evolutionary algorithms provide promising results dealing with the FS problem. In this paper, we proposed a correlation-aware FS approach based on the estimation of distribution algorithms (EDAs), which is categorized in population-based optimization methods. The EDA methods use an explicit probability distribution to generate new candidate solutions, and a sequence of probabilistic model updates is used to optimize the problem. The main contribution of our proposed method is considering both the importance of each feature alone and the interaction between features. To do this, we introduced a conditional probability scheme that considered the joint probability distribution of selecting two features. The other advantage of the proposed method is that it generates only two individuals in each iteration. However, the obtained results in the experiments proved that this does not weaken our approach to searching the entire space. The generated individuals compete in each iteration and evolve during the algorithm to find the best solution. Finally, we proposed a guiding mechanism that ensures that the number of features for each individual does not exceed the value determined by the chi-square distribution. This can directly increase the convergence speed of the algorithm due to the limitation on the number of features for each individual. The experimental results on synthetic and real-world datasets and the statistical analysis proved that the proposed method has been quite successful in both considering the correlated features and increasing the classification accuracy. As future work, it is interesting to extend our method to deal with multi-objective problems. Thus, we can benefit from the advantages of the proposed method in more real-world applications.

9.